\documentclass[runningheads]{llncs}

\usepackage{booktabs} 

\usepackage{amsmath}
\usepackage{url}
\usepackage{graphicx}
\usepackage{listings}
\usepackage{algorithm}
\usepackage{algorithmicx}
\usepackage{algpseudocode}
\newcommand{\topic}[1]{}

\begin{document}
\title{Design and Implementation of\\ Linked Planning Domain Definition Language}
\titlerunning{PDDLS}
\author{Michiaki Tatsubori\inst{1} \and Asim Munawar\inst{1}  \and  Takao Moriyama\inst{1} 
}
\institute{IBM Research - Tokyo\\
	\email{\{mich,asim,moriyama\}@jp.ibm.com}
}


\maketitle

\begin{abstract}
	Planning is a critical component of any artificial intelligence system that concerns the realization of strategies or action sequences typically for intelligent agents and autonomous robots. Given predefined parameterized actions, a planning service should accept a query with the goal and initial state to give a solution with a sequence of actions applied to environmental objects. This paper addresses the problem by providing a repository of actions generically applicable to various environmental objects based on Semantic Web technologies. Ontologies are used for asserting constraints in common sense as well as for resolving compatibilities between actions and states. Constraints are defined using Web standards such as SPARQL and SHACL to allow conditional predicates. We demonstrate the usefulness of the proposed planning domain description language with our robotics applications.
\end{abstract}

\pagestyle{plain}

\section{Introduction}
\label{sec:introduction}

Given a knowledge base and two logical instances $a$ and $b$, such as a peg and a hole, we often want to know if a role assertion axiom $(a, b):R$, such as the insertability of the peg into the hole, holds.
In practice, it is often useful if the knowledge base can come from multiple separate sources for the same domain.
Particularly, in addition to a "static" ontology database as the common sense in the domain, such as the general knowledge of the ``insertability'', we obtain a new piece of information in the environment on, for example, pegs and holes and their instant properties such as their shapes and sizes.
This kind of assumption of knowing unknowns in a knowledge base, also called the open-world assumption, characterizes Semantic Web technologies.



\topic{Making robot programming easy.}
Modular componentization of robotic cognition and action systems is the key for easy programming of cognitive robots or virtual agents~\cite{Munawar:ICRA2018:MaestROB}, which interact with humans to perform complicated tasks.
A robot may listen to an instruction from a human user, see pegs and holes on a table, and perform the sophisticated combination of pick, place, insert, and other actions to satisfy the instructed goal, such as ``Fill all the holes.''
A component called {\it{``planner''}} usually plays the orchestration role in robotic systems~\cite{Ghallab:2016:AutomatedPlanning} by accepting the goal and initial state as a problem and actions available as a domain to give a sequence of actions to solve the problem.
Cognitive solutions of visual recognition and natural language processing are becoming well componentized and are even available as off-the-shelf Web services (or APIs)~\cite{Waibel:2011:roboearth}\cite{Kehoe:ToASE2015:cloudroboticssurvey} to obtain such goals and initial states.

\topic{Skill repository does not scale with PDDL.}
Unfortunately, the componentization of actions in the sense of ``web-style'' have not yet been established to be readily available.
The ``web-style'' here indicates a system architecture with open standard-supported, loosely-coupled components.
Ideally, each type of action should be available on the Web with previously machine-trained or hand-configured models to control row-level robot actuations such as changing the joint angles of robot arms.
The exchangeable formats of machine-trained models, such as PMML and PFA~\footnote{\url{http://dmg.org/pmml/products.html}}, or the more recent neural model-dedicated ONNX~\footnote{\url{http://onnx.ai}} have been used or are becoming standardized.
However, though the planning domain definition language (PDDL)~\cite{McDermott:1998:PDDL,Fox:JAIR2003:PDDL21} is a de facto standard in artificial intelligence for providing commonly-understandable planning problems with very simplified metadata for actions, it is not designed to be used as metadata for indexing such models as actions for use in the Web architecture.

\begin{figure}[t]
	\centering
	\includegraphics[width=1.0\linewidth]{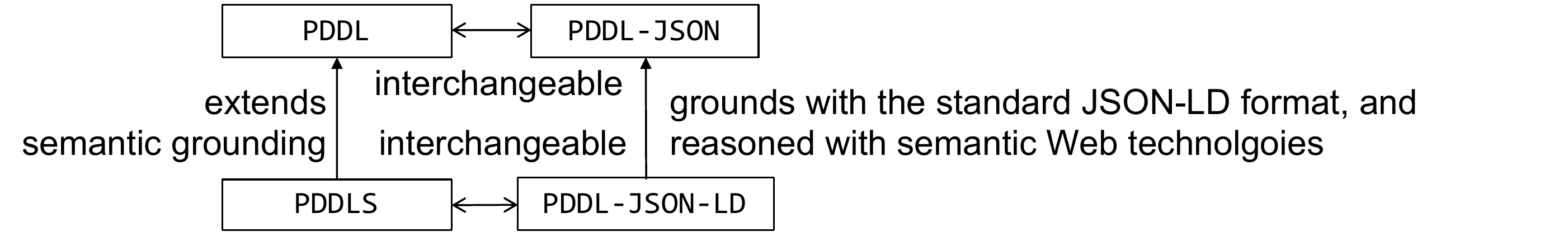}
	\caption{PDDLS's relationships to PDDL, JSON (PDDL-JSON) and JSON-LD (PDDL-JSON-LD).}
	\label{fig:pddl-jsonld}
\end{figure}

\topic{PDDL with semantics expect helps from Semantic Web technologies.}
This paper introduces the PDDLS (Planning Domain Definition Language with semantics) language and a reference design and implementation of its resolver.
The design goals of PDDLS are to:
\begin{itemize}
	\item minimize the syntactical extension of the base PDDL~\cite{McDermott:1998:PDDL,Fox:JAIR2003:PDDL21}, while
	\item helping planning domain descriptions to interoperate and to be reasoned leveraging Semantic Web technologies and assets~\cite{Horrocks:2008:OntoSemWeb}.
\end{itemize}
\noindent
While sharing the namespace concept found in Web-PDDL~\cite{Dou:2005:OntEngine} or PDDL/M~\cite{dornhege2012semantic}, PDDLS follows the practical annotation design of JSON-LD~\cite{Lanthaler:WSREST2012:JSONLD} to allow annotating any local symbols or terms to be bound to globally unique IDs as IRIs (internationalized resource identifier, defined in the RFC 3987 standard to extend the URI/URN/URL), which can then be resolved using Semantic Web~\cite{Shadbolt:2006:semanticweb} technologies.
Figure~\ref{fig:pddl-jsonld} depicts PDDLS's relationships to the base PDDL, the PDDL representation in JSON (PDDL-JSON), and the PDDLS representation in JSON-LD (PDDL-JSON-LD).
An additional reasoning step supported by Semantic Web technologies, if working well, would help a planner-like system to integrate actions developed by multiple vendors or communities.
In this research, we develop a knowledge-driven engine to give robots the power of processing the common sense knowledge that helps to better understand human commands.


The advantage of this approach is that we do not need to define grammar like ELI did to achieve a similar task ~\cite{connell2014extensible}.
As our approach uses ontology, we can update the knowledge at runtime, which makes extending the system to handle complex scenarios easy.
Even though here we considered only the orientation and the relationship of the objects to be able to put them on top of one another, our approach can be extended for any other required restrictions using ontology.

\topic{Help from OWL-style common sense is limited as we need to handle objects unknown at the time of KB creation.}
By using the knowledge and ontology of physical constraints and relationships, these abstractions allow the grounding of human instructions to actionable commands.
The framework performs symbolic reasoning at a higher level, which is important for long-term autonomy and making the whole system accountable for its actions.
Individual skills are allowed to use machine learning or rule-based systems. We provide a mechanism to extend the framework by developing new services or connecting it with other robot middleware and scripting languages.
This allows the higher level reasoning to be done using a PDDL (planning domain definition language) planner with the proposed extension, while the lower level skills can be executed as ROS (robotics operating system) nodes.


\topic{We demonstrate its effectiveness in a practical robotics use case.}
We show the capabilities of the framework by a scenario where a human teaches a task to a communication robot (Pepper) by demonstration.
The robot understands the tasks and collaborates with an industrial manipulator (UR5) to execute the task, using the action primitives that UR5 has previously acquired by learning or programming.
The industrial robot has the ability to perform physical manipulation, but it lacks the key sensors that can help in a particular situation (e.g. error recovery etc.).
In the scenario, the communication robot having these sensors can analyze and convert the plan to a command sequence for the robotic arm.
Semantic ontologies are used to resolve the plan.


This paper explores how semantic ontologies can help resolve semantically annotated problems as goals and initial states with semantically annotated actions with preconditions and effects.
As what we can do with simple OWL 2 semantics resolvers is quite limited, we extend the PDDLS resolver to provide description logics that are beyond the level of first-order description logics provided by OWL 2.
We also leverage some standard Semantic Web technologies: RDF, OWL, JSON-LD, and SHACL.

\topic{Contribution.}
The main contribution of this paper includes:
\begin{description}
	\item[PDDLS design and implementation] We extended PDDL leveraging the JSON-LD. A simple PDDLS-to-PDDL translator has been designed and implemented by leveraging semantic technologies.
	\item[PDDLS semantics] To overcome the limitations with OWL with typical descriptive logic (DL) resolvers, we extend the DL with a powerful role-construction operator. The extended resolver for the operator is formally defined.
	\item[Demonstration with a robotics use case.] PDDLS usage in an actual robotics scenario has been shown with a working cognitive robot system.
\end{description}

The rest of this paper is organized as follows.
We start with the section describing our motivation behind the work, and then we propose the PDDLS language with its syntax and semantics.
Then, we discuss semantic resolvers for PDDLS in terms of expressive power.
We demonstrate how it can be used in our actual robotics application in the following section.
After mentioning related work, we conclude the paper.
\section{Motivating Scenario -- Cognitive Robotics}
\label{sec:motivation}

\begin{figure}[t]
	\centering
	\includegraphics[width=0.53\linewidth]{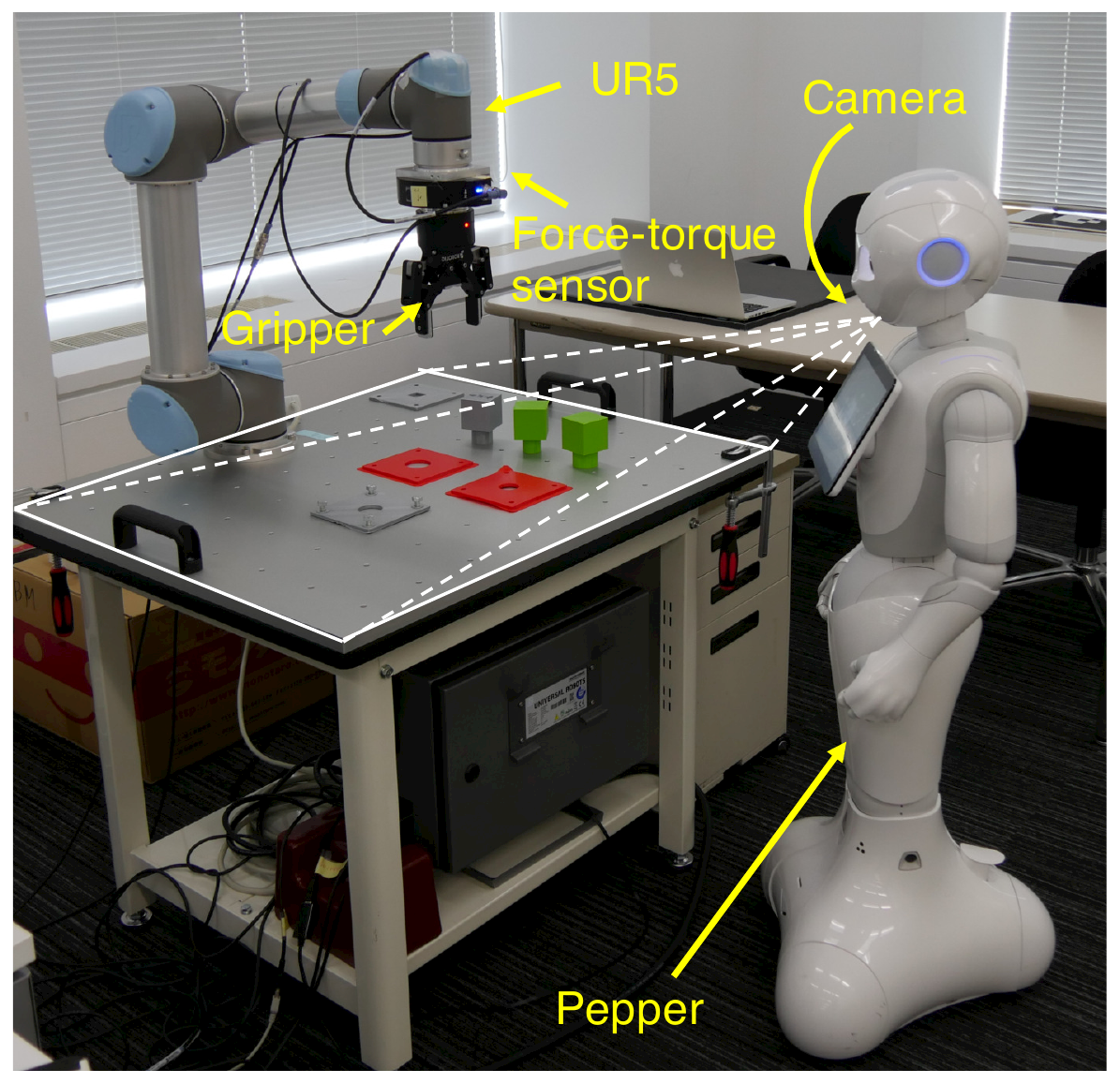}%
	\includegraphics[width=0.45\linewidth]{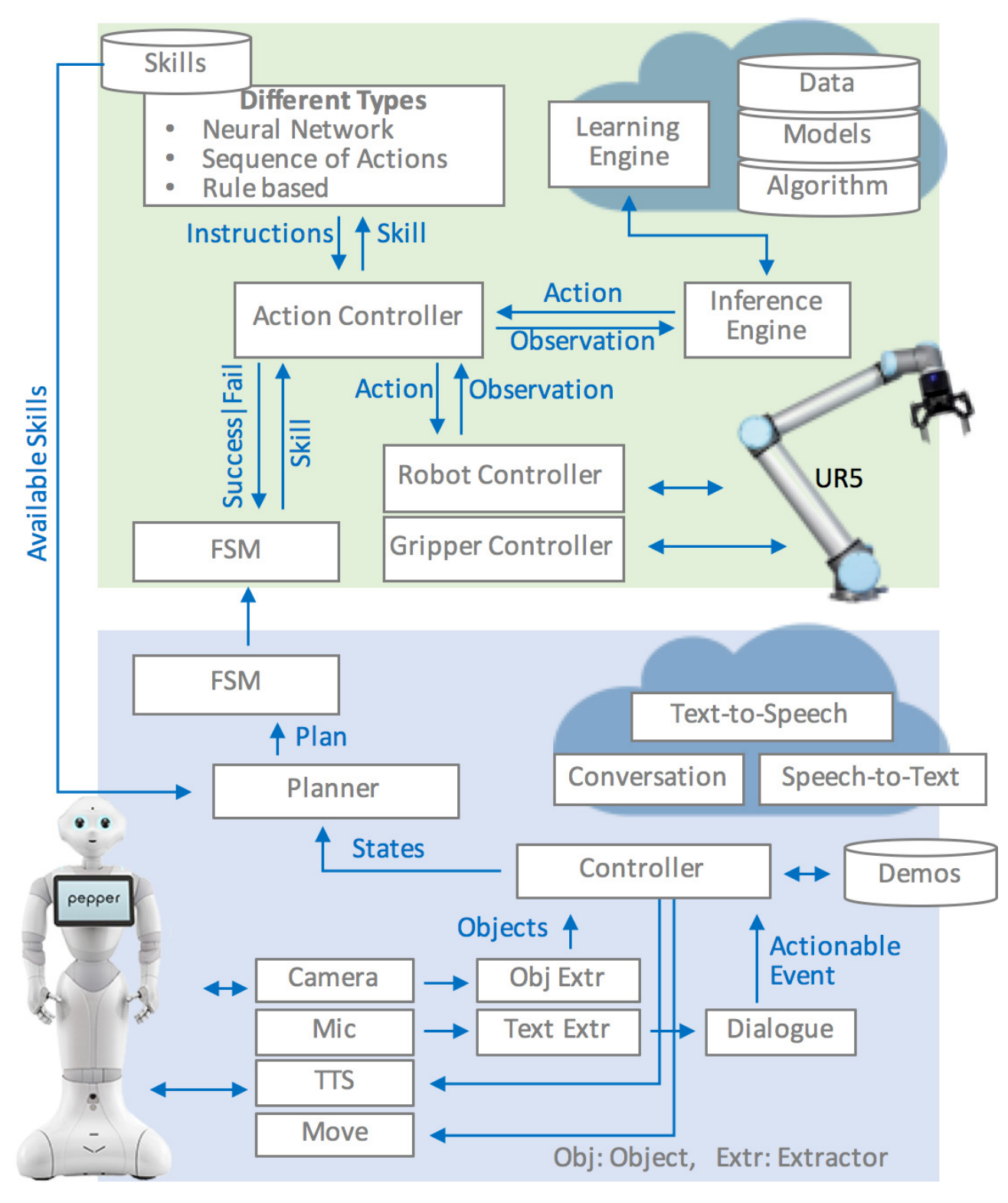}
	\caption{A scene picture (left) showing a motivating scenario and its system/service architecture behind. A cognition robot Pepper (right in the picture) and a actuator robot UR5 (left) work together.}
	\label{fig:setup}
\end{figure}

In the motivating scenario, two robots (UR5 and Pepper) collaborate with a human to perform a task.
Pepper is a humanoid robot from SoftBank that poses several sensors including vision.  It is mobile, but it lacks a gripper and cannot perform accurate physical manipulations.
On the other hand, UR5 (collaborative robot arm from Universal Robots) is a fixed industrial grade manipulator robot. UR5 can do physical manipulations with high precision and repeatability ($\pm 0.1$mm) but it moves blindly due to lack of any vision sensor.  Figure~\ref{fig:setup} shows a picture of two robots in the scenario and their system architecture.

\begin{figure}[t]
	\centering
	\includegraphics[width=0.70\linewidth]{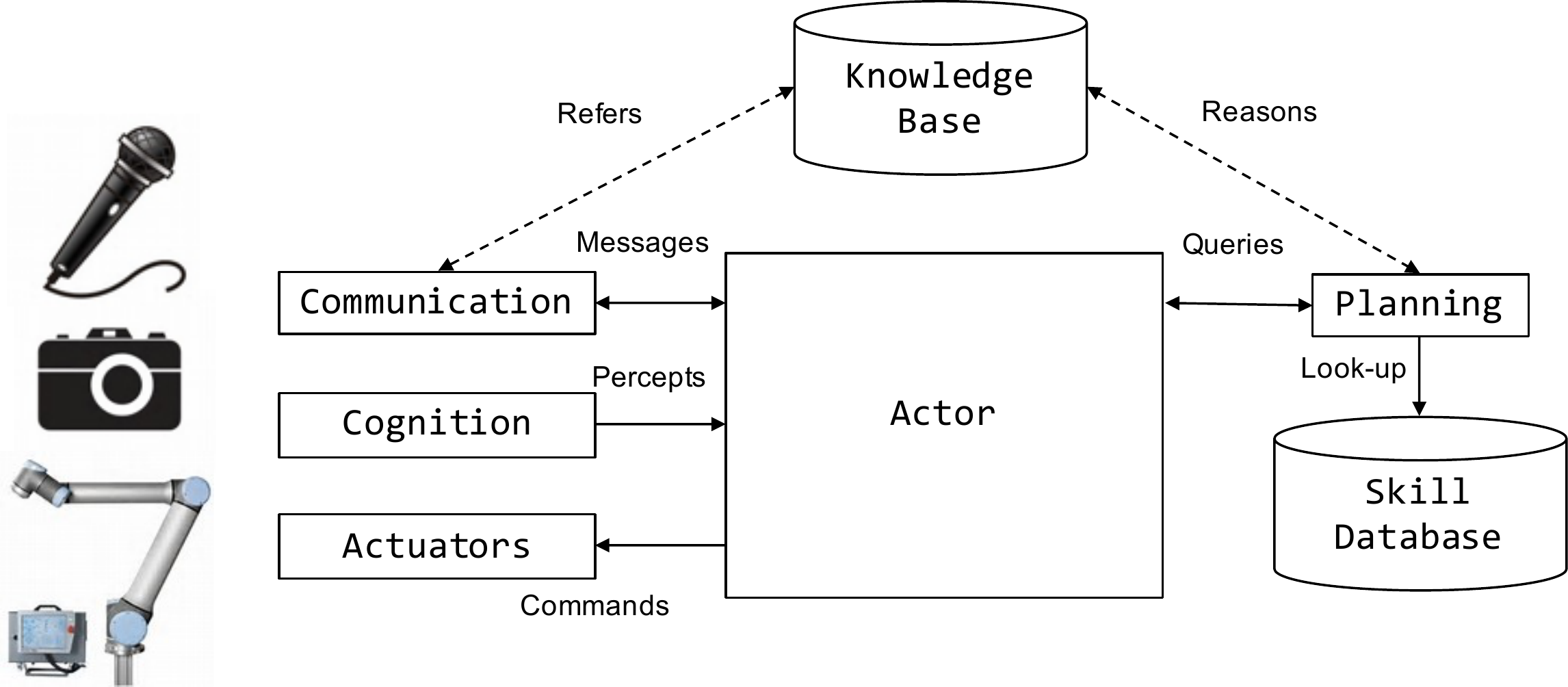}
	\includegraphics[width=0.80\linewidth]{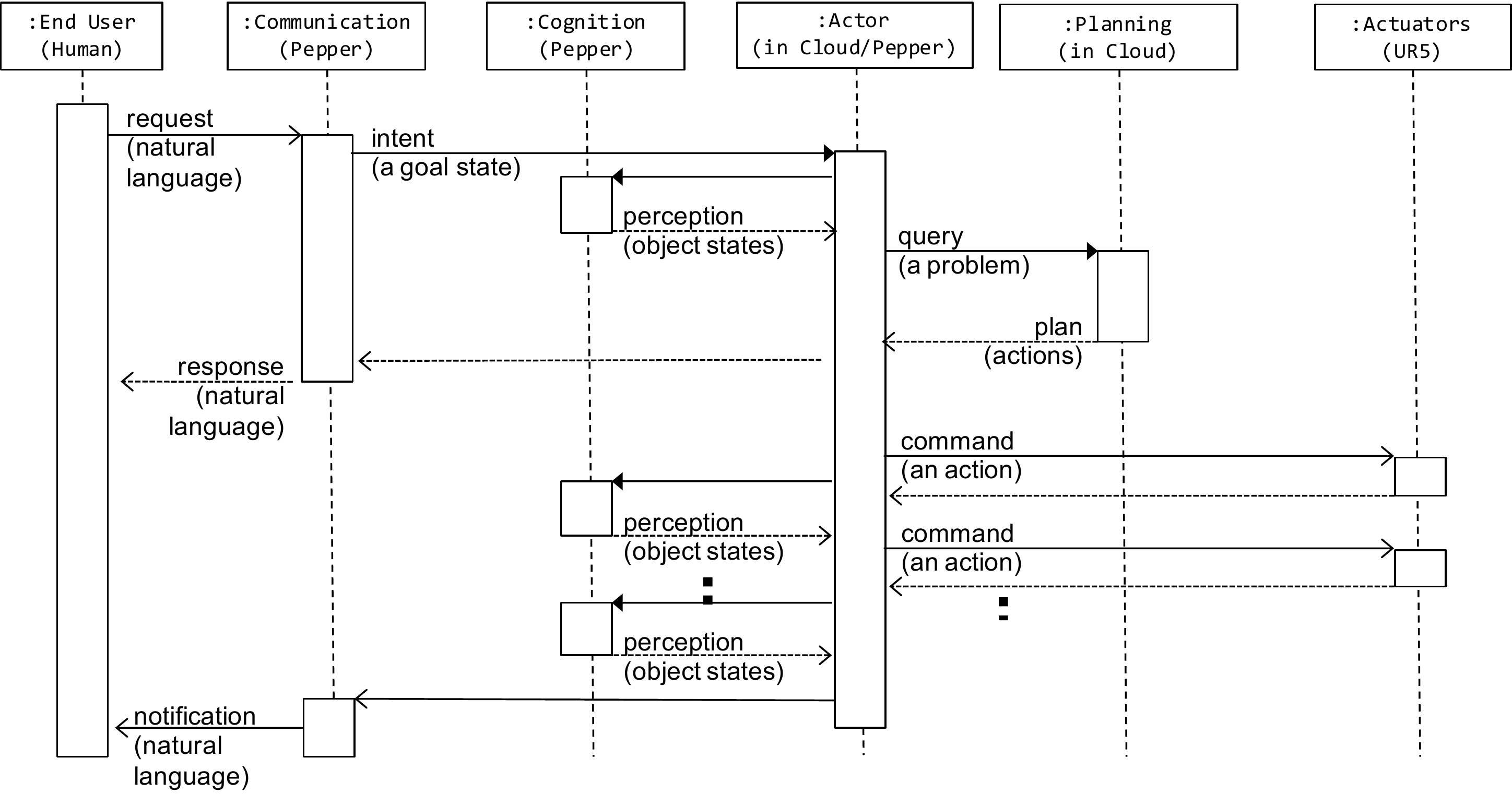}
	\caption{A conceptual view of an actor with planning, skill database and knowledge base components (upper) and a sequence diagram with planning (lower).}
	\label{fig:concept}
\end{figure}

While the actual system is complex as depicted in Figure~\ref{fig:setup}, the focus of this paper can be depicted as simple in Figure~\ref{fig:concept}.  The actual system involves various cognitive services in a Cloud (or components) for natural language processing (understanding/generation) and visual recognition (object identification and pose estimation).  The details of those services are out of scope of this paper.  Theoretically, it could be any kind of implementation or services.

In order to solve a large problem in a reasonable time, the actor in those conceptual views (Figure~\ref{fig:concept} )  should often employ a high-level planning component~\cite{Ghallab:2016:AutomatedPlanning}.  It performs planning using pre-specified symbolic state descriptors, and operators that describe the effects of actions on those descriptors.  For bridging the gap between a low level information and control, such as sensors and actuators for the physical environment, and high-level planning, recent work address the issues in constructing a symbolic description from natural language~\cite{Lu:IJCAI2017:ASPRobotPlanning}, from a continuous, low-level environment~\cite{Konidaris:AAAI2014:constructingsymbol}, and even from an abstraction hierarchy of skills~\cite{Konidaris:IJCAI2016:skillsymbolloop}, for use in planning. 

The scenario starts with a human having conversation with the Pepper robot to specify a task to complete.  Extracting the intent and entities of the order by using a Natural Language Understanding (NLU) service, the robot gets to know when to capture key frames to understand the current situation of objects and their properties in the environment, and with which situation the task is completed.

The initial and final frame are sent to a perception service that uses barcode pose detection to detect the location of all barcodes attached to objects. The barcode number of each part and the transformation between the barcodes and the objects are defined separately in the objects database.
The state of the final frame is determined by a relationship extractor through human demonstration, or simply by understanding from the conversation.  The domain as the available skills for the task is predefined and stored in a repository. Assume the goal state is computed to be (not (available hole)) for any $hole$ object.  In practice, we use a visual demonstration among human-robot conversation to infer such the goal state.

%
%
\section{PDDLS Syntax and Semantics}
\label{sec:pddls}

PDDLS has a minimal extension from PDDL to help planning domain descriptions to interoperate with each other and to incorporate externally-defined ``common sense'' through Semantic Web technologies.  Figure~\ref{fig:example-pddls} shows a normal PDDL domain description example except for the {\bf{:context}} declaration, which is of the syntactic extension from the basic PDDL.

\begin{figure}[t]
\scriptsize
\begin{lstlisting}
(define (domain example-ur5-domain)
  (:requirements :strips :adl :typing :semantics)
  (:|\textbf{context}|
    available - uri:cril/action/available
    insertable - uri:cril/action/insertable
    pick-n-insert - uri:cril/action/pick-n-insert )
  (:predicates
    (available ?object) ; pillar or hole is available
    (insertable ?piller ?hole) )
  (:action pick-n-insert
    :parameters (?pillar ?hole)
    :precondition (and
      (available ?piller)
      (available ?hole)
      (insertable ?piller ?hole))
    :effect (and
      (not (available ?piller))
      (not (available ?hole)) ) ) )
\end{lstlisting}
\caption{An example PDDLS domain description.}
\label{fig:example-pddls}
\end{figure}

The {\bf{:context}} declaration adds global namespaces for locally defined symbols in the description.  For this example, the symbol ``available'' is bound to a globally unique ID specified as an IRI (conceptually generalized URI or URL). 
Following the concept of JSON-LD~\footnote{https://www.w3.org/TR/2014/REC-json-ld-20140116/}, PDDLS allows PDDL to be interpreted as Linked Data, which provides a way to help the PDDL description interoperate at Web-scale.
The extended syntax is defined as partially shown in Figure~\ref{fig:syntax}.
It is primarily intended to be a way to use Linked Data in Web-scale robotics programming, to build interoperable planning services, and to store Linked robotics action skill metadata in repositories accessed by the planning services. 

\begin{figure}[t]
\scriptsize
\begin{verbatim}
<domain> = (define (domain <local name>)
                [<requirementsDef>]
                [<contextDef>]
                [<typesDef>] 
                [<constantsDef>]
                [<predicatesDef>]
                [<functionsDef>]
                [<constraints>]
                {<structureDef>} )
<contextDef> = (:context
                <URI> | { <termMapping> } )
<termMapping> = <term> - <URI>
\end{verbatim}
\caption{The part of PDDLS syntax in EBNF, only showing the part extended from the original PDDL.}
\label{fig:syntax}
\end{figure}

The specification of PDDLS context is designed to completely borrow from JSON-LD except for the syntactic difference for making it look PDDL.  While the syntax definition shown in this paper is a subset of JSON-LD 1.1 expressive power, it is straight forward to extend it to include the full JSON-LD expressive power, such as term expansion, type coercion, node properties, and so forth.  The current parser implementation does not cover all of those advanced language constructs yet.

The symbol-IRI mapping semantics is indirectly defined with our definition of translation from PDDLS to JSON-LD.  Once translated into JSON-LD, how symbols (or terms) in PDDLS are mapped to IRIs is defined according to JSON-LD specification.  Figure~\ref{fig:example-jsonld} shows how PDDLS in Figure~\ref{fig:example-pddls} is to be translated into JSON-LD.
See the first JSON array key,``@context'', as it is the special JSON-LD notation for annotating semantic context.
Its value is derived from the symbol-IRI mappings defined with the {\bf{:context}} declaration in the original PDDLS in Figure~\ref{fig:example-pddls}.

\begin{figure}
\scriptsize
\begin{lstlisting}
{
"|\textbf{@context}|": {
  "pddl": "uri:pddl",
  "available": "uri:cril/action/available", 
  "insertable": "uri:cril/action/insertable", 
  "pick-n-insert": "uri:cril/action/pick-n-insert"
}, 
"pddl:domain": "example-ur5-domain", 
"pddl:requirements": [":strips", ":adl", ":typing"],
"pddl:predicates": [
    {"clear": [{"?hole": null}]}, 
    {"available": [{"?pillar": null}]}, 
    {"insertable": [{"?pillar": null}, {"?hole": null}]}
  ], 
  "pddl:structure": [
    {"pddl:action": "pick-n-insert", ... }
  ]
}
\end{lstlisting}
\caption{The JSON-LD corresponding to Figure~\ref{fig:example-pddls}}
\label{fig:example-jsonld}
\end{figure}

A problem description in PDDLS, which can be regarded as a plan query over a skill repository, can be described similarly to the domain description explained above.  Figure~\ref{fig:example-pddls-problem} shows an example PDDLS problem.
Except for the {\bf{:context}} declaration, it is a normal PDDL problem description.

\begin{figure}[t]
	\scriptsize
\begin{lstlisting}
(define (problem example-problem)
  (:|\textbf{context}|
    available - uri:cril/action/available
    CylindricalPillar_1 - uri:cril/demo2/CylindricalPillar_1
    TriangularPillar_2 - uri:cril/demo2/TriangularPillar_2
    CylindricalHole_4 - uri:cril/demo2/CylindricalHole_4 )
  (:domain example-ur5-domain)
  (:objects
    CylindricalPillar_1
    TriangularPillar_2
    CylindricalHole_4 )
  (:init
    (available CylindricalPillar_1)
    (available TriangularPillar_2)
    (available CylindricalHole_4) )
  (:goal
    (not (available CylindricalHole_4)) )
)
\end{lstlisting}
	\caption{An example PDDLS problem (a query for a plan) description.}
	\label{fig:example-pddls-problem}
\end{figure}

\begin{figure}[t]
	\centering
	\includegraphics[width=0.49\linewidth]{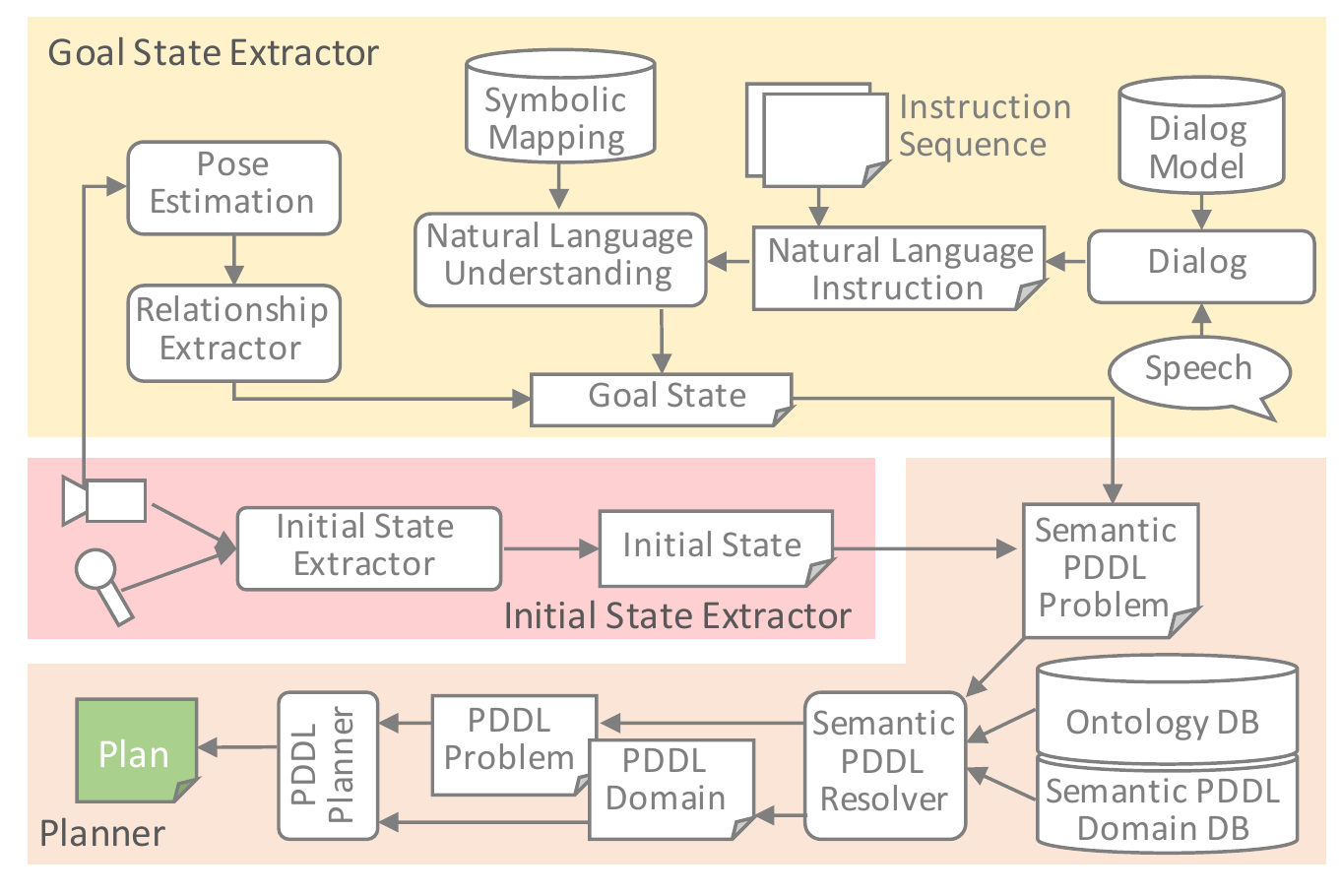}
	\caption{Process flow with a normal PDDL planner (solver).  PDDLS resolver (the bottom righter box) produces PDDL problem and domain for a normal PDDL planner (the bottom lefter box) to generate a plan.}
	\label{fig:planner}
\end{figure}

The actual interpretation of such semantic annotations in PDDLS is off-loaded to a separate semantic reasoning system, which is discussed in the following section, to keep the PDDL language constructs definitions simple.
Semantic reasoners may append arbitrary PDDL constructs into the base PDDL description.
After semantic reasoning, action and predicate symbols reasoned to be mapped to a same IRI in PDDLS are renamed to be an unified name with in resulting PDDL descriptions.
Otherwise, as shown in Figure~\ref{fig:planner}, PDDL descriptions are recovered and given to a normal PDDL planner, which produces a sequence of action skills with actual object arguments.

\section{Semantic Reasoning for Role Construction}
\label{sec:semantics}

McGuiness and Noy \cite{noy2001ontology} provide five reasons to share common understanding of the structure of information among people or robots, to enable reuse of domain knowledge, to make domain assumptions explicit, to separate domain knowledge from the operational knowledge, and to analyses domain knowledge for the development of an ontology.
Until now, many research efforts have been performed in order to apply Semantic Web technologies in various engineering fields for intelligent system development~\cite{kunze2010putting,lam2012evaluation,skulkittiyut2014building,tenorth2010understanding}.
Once PDDL is integrated with Semantic Web, it could potentially benefit from such the ontologies.

How much the``common sense'' in an ontology DB can potentially help PDDLS resolution, in other words, the expressive power of the ontology, is up to reasoning.  In terms of PDDL, the predicative axioms such as ``insertable'' should be inferred from the common sense knowledge base.  For example with Figure~\ref{fig:example-pddls-problem}, be aware that the given initial state misses:

\begin{verbatim}
  (insertable CylindricalPillar_1 CylindricalHole_4)
\end{verbatim}

\noindent
Without this axiom proposition, the action {\bf{pick-n-insert}} in Figure~\ref{fig:example-pddls} cannot be applied to these pillar and hole as the action requires it as a part of its precondition.  A common sense knowledge base is desired to be capable of providing such a logic.

\subsection*{OWL-based Knowledge Base}

Before considering complicated Robotics programming situations where ontologies should help, let us see simpler examples where ontologies can help.  We illustrates how A-box and T-box of a knowledge base (KB) contributes in easing/enabling robotics programming.

The human input command given in natural language is parsed using a Natural Language Understanding (NLU) service, which identifies the action, subject and object.
For example, ``Put the box on the book'' would return the semantic roles as ``Put (Action) the box (subject) on the book (object)'' using JSON.
It also identifies the keywords which helps to understand the objects.
For example: book and box are the keywords detected from the above example.

\topic{Sensor Data}
Our Knowledge Engine receives data from the robot sensors. All of the objects are registered in the model database which includes the object type information.
For example, whether the object is a box, book, pen or a glass.
All the objects are an instance of the object ontology, which has been explained in the next section.
For each objects we have two types of information: environment dependent and object dependent information.
Environment dependent information contains the object position and orientation in (X, Y, Z) format. In the model database, object dependent information includes the size of the object, weight, center of gravity, hardness and color information.

\topic{Check Validity and Robot Actions}
Basic robot actions are defined and it can understand basic commands. Each basic command can be mapped to a defined action. For example, our robot system can understand the command ``Put A B''. The robot will pick the object A and put that on top of object B.

In our knowledge engine, robot has a model database and in front object list from its sensors. From that list, it will able to find the subject and object positions. 
Knowledge described using  OWL can help robots to understand whether a command is meaningful to process. Let's assume that o1 is-an Object and that  o2 is-a Pyramid.  Any object cannot-be-on-top-of any object in SharpHeadObject (ABox). Pyramid is-subclass-of SharpHeadObject. (TBox), because of that o1 cannot-be-on-top-of o2. 

From the model database it can find the object type and from the ontology the robot can check whether the subject can be put in top of the object. If it is not valid then robot will inform the user that the command cannot be performed. If it is valid then, the robot will able to perform the put operation. 

For example, ``Put the book on the box'' can be performed, but ``Put the book on the Pyramid'' cannot be performed. Our knowledge engine can check the object type in run time and create an instance of that object in our ontology. After that using HermiT reasoner it can validate whether the object can be put on top of another or not.

This can be used in processing a command like ``Stack everything''. The robot can determine which objects can be on top of what, using this ``put'' function. For example, if there is a box, a book and a pyramid then the pyramid should be on top then book or box.

\subsection*{Conditional Predicate beyond OWL}

Unfortunately, the ontology defined for a skill, such as the {\bf{insertable}}  predicate for Figure~\ref{fig:example-pddls}, is often not known when specifying the world state through general cognitive services.  Recognition of the current situation is sometimes performed regardless the requirements of skills in a repository.  To be practical, a modular recognition service should be independent from a number of various skills.  Instead, the recognition service should just focus on the primitive states such as shapes and sizes of objects in the scene.

Predicates for unknown objects should be inferred through the primitive object ontology and the ``common sense'' ontology, which can be separately defined both from the object ontology and the skill ontology.  We call such predicates {\it conditional}, as its assertion is conditional according to the reasoning of object properties.  In addition to role-inclusion and role-equivalence axioms supported by DLs, now we need role-construction axioms.


A working example ontology with a conditional predicate {\bf{insertable}} looks like as follows in RDF/Turtle:
%
\begin{verbatim}
ex_action:insertable
 pddls:establishedWith ex_shapes:InsertableConstraint ;
 pddls:establishedWith ''.. SPARQL here .."@sparql .
\end{verbatim}
%
\noindent
The {\bf{establishedWith}} is the special predicate introduced for the role-constructive reasoning.  Objects of the axiom can be either a string literal representing SPARQL~\footnote{https://www.w3.org/TR/sparql11-query/} query or a SHACL~\footnote{https://www.w3.org/TR/shacl/} constraint expression.

\subsection*{Conditional Predicate Semantics}

Let $\mathcal{I}$ be an interpretation, which consists of a set $\Delta^\mathcal{I}$, called the domain of  $\mathcal{I}$, and an interpretation function $\cdot^\mathcal{I}$.
The function  $\cdot^\mathcal{I}$ maps
each atomic concept $A$ to a set $A^\mathcal{I} \subseteq \Delta^\mathcal{I}$,
each atomic role $R$ to a binary relation set $R ^\mathcal{I} \subseteq \Delta^\mathcal{I} \times \Delta^\mathcal{I}$, and
each individual name $a$ to an element $a^\mathcal{I} \in \Delta^\mathcal{I}$.

In addition to those supported with the $\mathcal{SPOIQ}$ DL for OWL 2, $\mathcal{I}_\text{OWL}$,
we introduce an additional interpretation $\mathcal{I}_\text{CP}$ with a constructor $\textit{Establish}$, defined as a syntax of:

$ \textit{Establish}(R, q)$

\noindent
with its semantics of:

$\{ R(a, b) | (a, b) \in  {[[q]]}_{\Delta}  \}$

\noindent
where the evaluation of a graph pattern $q$ over an RDF dataset $\Delta$ is denoted by $[[q]]_{\Delta}$.  We follow the approach in \cite{Perez:ToDS2009:SPARQL,Gutierrez:PoDS2004:semanticweb} defining the query evaluation semantics as the set of mappings that matches the dataset $\Delta$.   Intuitively speaking, the individual $q \in \Delta$ represents a query function to test any ordered pair (2-tuple) of individuals in $\Delta$ to be role-asserted with a role $\mathcal{R}$.
Note that the semantics is defined using $\Delta$ rather than $\Delta^\mathcal{I}$ to avoid the complication with the higher-order construction.

For the extended interpretation $\mathcal{I}_\text{OWL+CP}$ is then defined with an interpretation function:

\begin{equation}
\cdot^{\mathcal{I}_\text{OWL+CP}}
=
((\cdot^{\mathcal{I}_\text{OWL}})^{\mathcal{I}_\text{CP}})^{\mathcal{I}_\text{OWL}}
\end{equation}
\noindent
Note that this is very simplified implementation restricting the recursive application of interpretation $\mathcal{I}_\text{CP}$ to avoid undecidability of reasoning.



%
%
%
%

\subsection*{Implementation}
\label{ssec:implementation}

In logic, the term decidable refers to the decision problem, the question of the existence of an effective method for determining membership in a set of formulas, or, more precisely, an algorithm that can and will return a boolean true or false value that is correct (instead of looping indefinitely, crashing, returning "don't know" or returning a wrong answer). Logical systems such as propositional logic are decidable if membership in their set of logically valid formulas (or theorems) can be effectively determined.

A theory (set of sentences closed under logical consequence) in a fixed logical system is decidable if there is an effective method for determining whether arbitrary formulas are included in the theory. Many important problems are undecidable, that is, it has been proven that no effective method for determining membership (returning a correct answer after finite, though possibly very long, time in all cases) can exist for them.
Many reasoners use first-order predicate logic to perform reasoning; inference commonly proceeds by forward chaining and backward chaining.






\section{Application Examples with Cognitive Robotics}
\label{sec:example}

\begin{figure*}[t]
	\centering
	\includegraphics[width=0.98\linewidth]{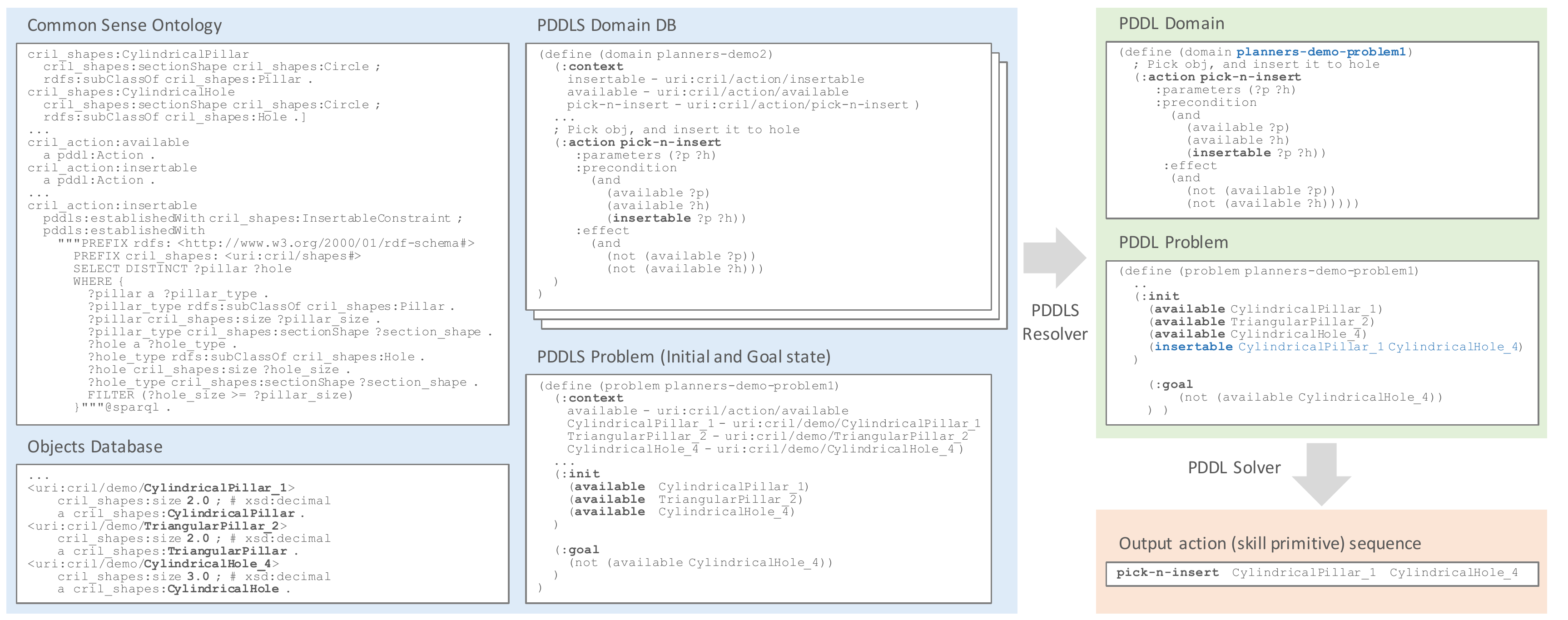}
	\caption{Input and output of PDDLS Resolver and PDDL Solver.}
	\label{fig:flow}
\end{figure*}

In order to demonstrate the usefulness of a common sense ontology built using the proposed role-construction operator, we show a working application with coginitive robotics.
Based on the low-level commands available to execute with UR5, such as forceControl(Fx, Fy), grasp(state), moveTo(x, y, z, Rx, Ry, Rz), we assume a higher level abstraction called skill primitives are provided for UR5.  We define skill as a piece of logic that can consume sensor input and generate low-level commands. A skill is an atomic operation that performs a part of the overall task. Skills however cannot be executed on their own.
It can be implemented as either a list of low-level commands, a set of rules, or a machine-learned model.
In addition to several simple skills that UR5 can perform, the insertion skill uses machine learning. It was trained by reinforcement learning using a learning service in a cloud. The training method for the insertion skill is very similar to T. Inoue et al.~\cite{conf:iros:inoue2017}.

Fig.~\ref{fig:flow} shows a simple example to illustrate how the system works.
First, we have Common Sense Ontology, which contains various kind of common sense which can not be acquired by cognition process.
Cylindrical is a Peg with it's section type circle, in the other hand, Triangular Peg has triangular section shape.
Cylindrical Hole is a Hole with its section (or hole) shape circle.

The Common Sense Ontology also contains a notion of ``insertable'' described in SPARQL format; if object A and B are type of peg and hole, respectively, with same  section shape, and the section size of A is equal to or smaller than that of object B, A is insertable to object B.  Alternative representation in SHACL is also available.

Second, PDDLS Domain DB contains a set of skills defined in the system. Each of skill is describe as PDDL augmented by semantics.
In this example, ``pick-n-insert'' is defined with its pre-conditions and effects.
All symbols are bound to be globally identifiable references as URIs.

The PDDLS resolution process starts when the Cognition System in the Fig. \ref{fig:concept}. captures environment information and goal.
Environment information is stored into the Object Database, which now includes a Cylindrical Peg, a Triangular Peg, and a Cylindrical Hole.
Assume we have required to fill the Cylindrical Hole with something.
We generate PDDLS Problem file, which contains a list of object available in the environment, and a goal described as q{(not (available CylindricalHole\_4))} meaning the hole is filled with something.
Also included in the PDDLS Problem file is a context information to define required relationships (``available'' in this case).

The PDDLS resolver then generates a runnable PDDL problem and domain file, using the semantic annotations to resolve necessary constraints.
These constraints are shown in blue color in the figure and they are required by the PDDL solver (PDDL planner) to find a valid solution.
The problem and the domain files are used by the PDDL solver to output the correct sequence of actions, which in this case is to perform the ``pick-n-insert'' to put the cylindrical peg in the cylindrical hole.

\begin{figure*}[t]
	\centering
	\includegraphics[width=0.98\linewidth]{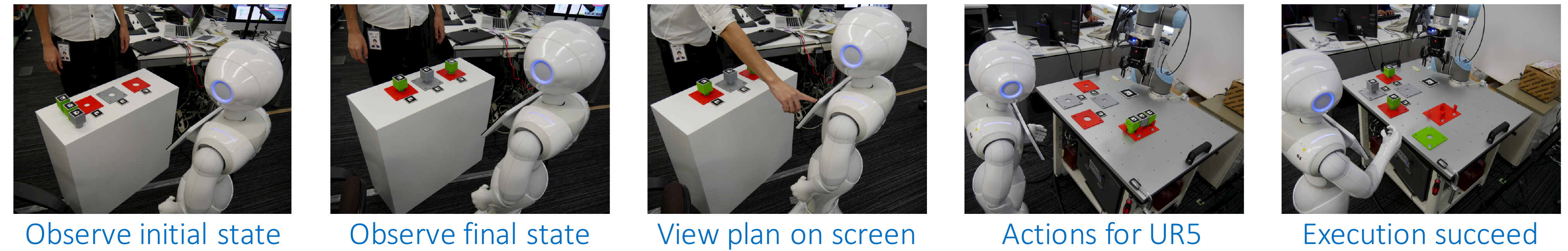}
	\caption{Demonstration with the motivating scenario.}
	\label{fig:movie}
\end{figure*}

Given the plan resolved the robots work as pictured in Figure~\ref{fig:movie}.  The first two pictures show the recognition of the initial state and the final (goal) state by Pepper.  Given those new information, the planning system can generate a plan to confirm with a human instructor.  The last two pictures demonstrate that planned actions are actually taken by UR5 and successfully finished.

\section{Related Work}
\label{sec:relatedwork}

Extending Description Logics, which underpins the Web Ontology Language OWL,　is one of the royal roads in terms of logic research.  Recent work has shown that many rule languages for Semantic Web, such as SWRL, can be expressed within the Description Logic paradigm, thus within the OWL language.~\cite{Carral:ESWC2012:DLRules}.  It can avoid undesirable properties such as undecidability.  Unfortunately, it does not seem to applicable to much more powerful descriptive operators such as what we introduced in this paper.  The rest of paper discusses other related work.

\subsection*{Common Sense Knowledge Base}

The idea of building a common sense knowledge base is very old as found in studies of the symbolic artificial intelligence paradigm from the mid-1950s until the late 1980s~\cite{Haugeland:1989:VeryIdea}, which are based on high-level "symbolic" (human-readable) representations of problems, logic and search.
The most successful form of symbolic AI is expert systems, which use a network of production rules.

Cyc~\cite{Lenat:CACM1995:CYC} is among  the most famous projects attempting to assemble a comprehensive ontology and knowledge base that spans the basic concepts and "rules of thumb" about how the world works.  One of its goals is to enable AI applications to perform human-like reasoning and be less "brittle" when confronted with novel situations that were not preconceived.  For Cyc ontologies and the global common sense knowledge base, CycL~\cite{Lenat:1991:CycL} is used as the ontology representation language.

Among very powerful CycL expressive features including reflection, implication rules is a the key expression power for the so-called core/domain theories in the Cyc common sense knowledge base.   What we are proposing in this paper might be considered as a theoretical subset of such the feature.   Our focus is to better align with Semantic Web technologies in terms of Web standards.




\subsection*{Semantic Web for Robotics}
\label{ssec:standards}

%

Our work is not the first to introduce semantic technologies in robotics as knowledge representation plays a fundamental role in the integration of planning in robotics.
A prominent example is the KnowRob system~\cite{Tenorth:IROS2009:KnowRob}, which combines knowledge representation and reasoning methods for acquiring and grounding knowledge in physical systems, accessing ontologies represented in OWL.

KnowRob2 is one of the most advanced knowledge processing systems for robots that has enabled them to successfully perform complex manipulation tasks~\cite{Beetz+:ICRA2018:KnowRob2}.
It is an extension and partial redesign of KnowRob~\cite{Tenorth+Beetz:2013:KnowRob}.  One of redesign is the provision of an interface layer that unifies very heterogeneous representations through a uniform entity-centered logic-based knowledge query and retrieval language. KnowRob2 is designed to leverage concepts and results from motor cognition and robot control to extend AI reasoning into the motion level details, rather than incorporating constraints from common sense. 

The work done by Lu et al. shares the same goal as our application example to be shown later in this paper.  They propose a novel integrated task planning system for service robots in domestic domains~\cite{Lu:IJCAI2017:ASPRobotPlanning}.  Their focus is on natural language understanding using FrameNet.  While we choose PDDL as the intermediate language, they use Answer Set Programming, which is a task planning framework with both representation language and solvers.

Erdem et. al. presented  a  formal  framework  that  combines high-level  representation  and  causality-based  reasoning  with low-level geometric reasoning and motion planning ~\cite{erdem2011combining}. They use a  geometric reasoner  which guides  the  causal  reasoner  to  find  feasible  kinematic solutions, which is different from our approach.
%
%
Gaschler et al. combine the power of symbolic, knowledge-level  AI  planning  with  the  efficient  computation  of volumes,  which  serve  as  an  intermediate  representation  for both  robot  action  and  perception ~\cite{gaschler2013kvp}.
\section{Concluding Remarks}
\label{sec:conclusion}

This paper proposed PDDLS, a minimal extension to PDDL, for addressing the problem in providing a repository of actions generically applicable to various environmental objects, based on semantic web technologies.
Constraints are defined using Web standards such as SPARQL and SHACL to allow conditional predicates.
With it, ontologies can be used for asserting constraints in common sense as well as for resolving compatibilities between actions and states.
We demonstrated its usefulness with a robotics application.

\bibliographystyle{splncs04}
\scriptsize
\bibliography{bibs/pddl,bibs/ontology,bibs/robotics,bibs/cril,bibs/graphdb} 

\end{document}